\documentclass[runningheads]{llncs}

\usepackage{eccv}

\usepackage{eccvabbrv}

\usepackage{graphicx}
\usepackage{booktabs}
\usepackage{wrapfig}
\usepackage[misc]{ifsym}
\usepackage{tablefootnote}
\usepackage[table]{xcolor} 
\usepackage{nicefrac} 
\usepackage{amsfonts}
\usepackage{booktabs}
\usepackage{url}
\usepackage{algorithm}      % For algorithm environment
\usepackage[noend]{algpseudocode} 

\usepackage[accsupp]{axessibility}  % Improves PDF readability for those with disabilities.

\usepackage{hyperref}

\usepackage{orcidlink}

\begin{document}

\title{SynVAR: Synergizing Spatial and Semantic Alignment in Visual Autoregressive Model} 

\titlerunning{SynVAR}

\author{Zhennan Chen \inst{1*} \and
Tianxing Shi \inst{1*}  \and
Pengcheng Xu \inst{2} \and  Kepan Nan \inst{1} \and \\
Qian Wang \inst{3} \and
Zili Yi\inst{1} \and
Jian Yang\inst{1} \and Ying Tai\inst{1}$^\dagger$}

\authorrunning{Z. Chen et al.}

\institute{State Key Laboratory for Novel Software Technology, Nanjing University \and
Western University \and
JIUTIAN Research, CMCC, China \\
\url{https://github.com/NJU-PCALab/SynVAR}}

\maketitle
\begingroup
\renewcommand\thefootnote{}
\footnotetext{$^\star$Equal Contribution. \\ $^\dagger$Corresponding Author.}
\endgroup
% \vspace{-4mm}
\begin{figure*}[!h]
    \centering
    \includegraphics[width=\textwidth]{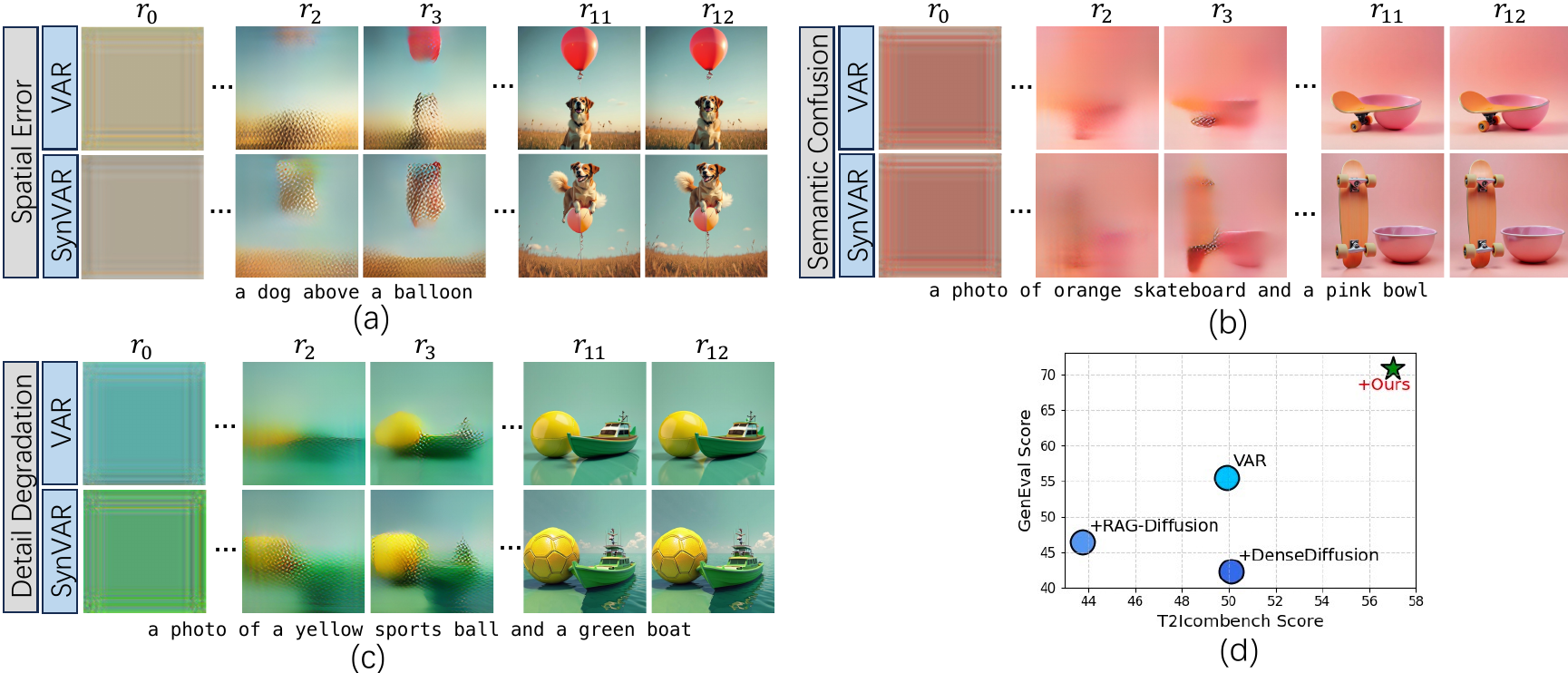}
    % \vspace{-6mm}
    \caption{(a)(b)(c): In complex scenes, existing VAR models are prone to spatial error, semantic confusion, and detail degradation. (d) Due to different formulations and modeling of the image, the solutions in existing diffusion models cannot play a positive role, while our method can significantly enhance the generation performance.
    }
    \label{Fig:teaser}
    \vspace{-8mm}
\end{figure*}

\begin{abstract}
VAR has gained widespread popularity due to its next-scale prediction paradigm. However, it faces substantial performance bottlenecks when handling complex scenes with multiple objects and attributes. Existing diffusion-based enhancement methods fail to adequately address the unique challenge of cross-scale error propagation and accumulation in VAR. To this end, we propose SynVAR, the first training-free enhancement framework specifically tailored for the VAR paradigm, which introduces a spatial-semantic collaborative control strategy to effectively suppress propagation error and improve generation quality. SynVAR comprises three key components: (1) Global guidance to ensure reasonable spatial structure in the early stages, (2) Receptive field constraints to mitigate early-stage semantic confusion, (3) High-frequency compensation to recover fine-grained details. Extensive quantitative and qualitative experiments demonstrate the significant improvements in the ability of SynVAR to enhance the VAR’s capability for complex scene modeling.
\end{abstract}
% \vspace{-6mm}

\section{Introduction}
\label{sec:intro}

The Visual Autoregressive (VAR) model \cite{tian2024visual} revolutionizes the generation paradigm of “next-token prediction” in the previous AR models with “next-scale prediction”. This new formulation significantly advanced generated images' quality and sampling efficiency and induced new AR-based large-scale text-to-image models such as Infinity \cite{han2024infinity}, Switti \cite{voronov2024switti}, and VARGPT \cite{zhuang2025vargpt}. Though thriving in quality and efficiency, they show deficiencies in spatial and semantic correctness in complex scenes, especially involving multiple objects, attributes, and spatial relations, which is essential and crucial for high-quality image generation. We demonstrate these issues in Figure~\ref{Fig:teaser}(a)(b)(c). %\textcolor{red}{several sentences introduce semantic confusion, etc.}
Specifically, the VAR may not accurately control the spatial relation of objects, and demonstrate visual semantics, and imagery details according to the text guidance in complex scenes. These indicate that the spatial and semantic relations in the VAR are not well aligned.
To address these, we investigate the pattern within the VAR transformer during generation and propose a train-free framework for more precise and controllable generation of complex scenes, enhancing the spatial and semantic alignment in the VAR paradigm.

\begin{figure*}[h]
    \centering
    \includegraphics[width=\textwidth]{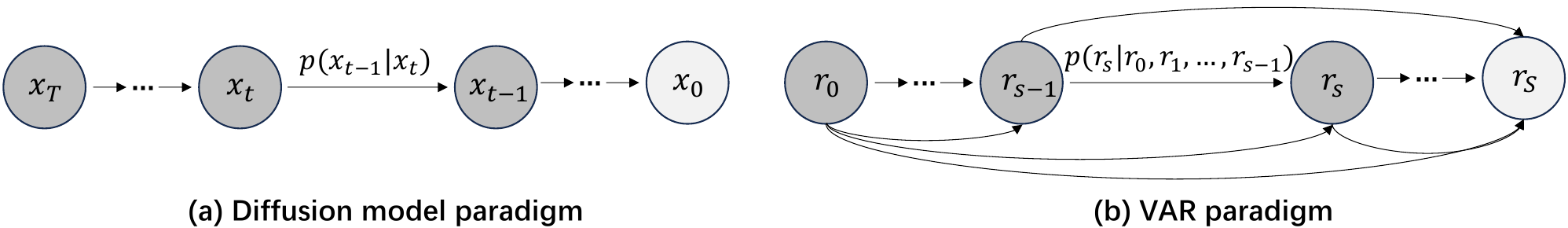}
    % \vspace{-6mm}
    \caption{Graphical models for diffusion and VAR inference models. Compared with the diffusion model, VAR is more prone to error accumulation and propagation.}
    \label{Fig:paradigm_difference}
    % \vspace{-4mm}
\end{figure*}

In our investigation, simply extending diffusion-based enhancement methods such as iterative refinement \cite{chen2024region, bar2023multidiffusion} and local attention \cite{kim2023dense,yang2024mastering} to VAR faces a significant degradation, as shown in Figure~\ref{Fig:teaser}(d). This is because VAR uses the different formulation and modeling of images from diffusion. As shown in Figure \ref{Fig:paradigm_difference}, in VAR, the prediction of each image scale is conditioned on all previous scales, whereas the image latent in diffusion is only conditioned on that previous one. Such a difference causes the spatial and semantic errors at the low-resolution scale of VAR more prone to propagate and accumulate and difficult to restore fine texture details.
%As shown in Figure \ref{Fig:teaser}(b), 
This characteristic of VAR amplifies the degradation of VAR’s spatial and semantic misalignment in complex scenes, which invalidates the techniques to improve the image quality. Thus, this motivates us to rethink spatial and semantic relations and enhance the generation rationality of VAR in complex scenes.

To address the issues above, we propose a collaborative correction mechanism for spatial and semantic control termed as SynVAR, which to our knowledge is the first training-free framework tailored for text-to-image models under the VAR paradigm. Specifically, the framework improves the spatial and semantic control from three aspects: 1) The global guidance optimizes the spatial partitioning strategy at the low-resolution scale stage and introduces explicit spatial priors in the cross-attention mechanism, ensuring the structural integrity of the image. 2) The receptive field constraint dynamically adjusts the interaction range of the front self-attention layers, effectively suppressing semantic confusion. 3) The high-frequency compensation enhances the high-frequency components of the features, enhancing the detail generation. Global guidance and receptive field constraint align spatial and semantic elements effectively, preventing error accumulation and ensuring that the final image accurately reflects the desired content. On this basis, the high-frequency compensation mechanism further enhances the detail performance of the image. Experimental results show that SynVAR consistently improves performance on various models such as Infinity and Switti, achieving average improvements of 19.6\% and 7.9\% on the overall score of the Geneval \cite{ghosh2023geneval} and T2I-CompBench \cite{huang2023t2i}, respectively. 

Our contributions are summarized as follows:
\begin{itemize}
    \item  We conducted a detailed investigation into the failure mechanisms of the VAR model in complex scenes. Based on these findings, we introduce SynVAR, the first training-free enhancement framework designed specifically to address these issues in the VAR paradigm.
    \item We propose an innovative approach for spatial and semantic alignment that combines global spatial guidance, receptive field constraints, and high-frequency compensation. This collaborative mechanism effectively reduces decision errors during the generation process and prevents the propagation of early-stage mistakes.
    \item As a plug-and-play solution, SynVAR significantly improves the performance of various VAR-based text-to-image models on different benchmarks, proving its effectiveness.
\end{itemize}

% \vspace{-4mm}

\section{Related Work}
\noindent{\textbf{Visual Autoregressive Modeling.}}
Inspired by the success of language models \cite{achiam2023gpt,brown2020language,touvron2023llama,touvron2023llama2,sun2024autoregressive,liu2024lumina} in the visual generation domain, a discrete representation-based autoregressive paradigm has gradually been established. Early efforts,  such as VQ-VAE \cite{van2017neural} and VQGAN \cite{esser2021taming}, established a strong foundation for visual autoregressive modeling by introducing the concept of encoding images into discrete visual tokens. These methods enabled the generation of images through a sequence of discrete symbols, similar to how text is processed in language models. Subsequent efforts \cite{chang2022maskgit, chang2023muse, razavi2019generating,deng2024autoregressive} introduced this paradigm into text-to-image synthesis. Building on these developments, VAR proposed the "next-scale prediction" strategy, transforming traditional one-dimensional token prediction into multi-scale image block prediction, thereby significantly enhancing both generation quality and semantic consistency. Recent advances such as Infinity \cite{han2024infinity}, Switti \cite{voronov2024switti}, and STAR \cite{ma2024star} have further optimized image detail representation and text-image alignment from different perspectives. VARGPT \cite{zhuang2025vargpt} also adopts the next-scale prediction paradigm for visual autoregressive generation, as part of broader multimodal visual understanding and generation efforts. Despite these successes in standard scenarios, existing VAR models still struggle with challenges in spatial and semantic alignment when handling complex prompts involving multiple objects and intricate attribute combinations.
% \vspace{-9mm}

\noindent{\textbf{Complex Scene Generation.}}
With the development of diffusion model \cite{ho2020denoising,song2020denoising,chen2023diffusion,peebles2023scalable,ramesh2022hierarchical,saharia2022photorealistic,nan2024openvid, zhou2024migc, zhou20243dis, chen2025dip, yang2025survey,chen2026l2p,yang2024d,yang2025omnivton,zhao2026luve,zhao2026zero,zhao2024wavelet,zhao2026learning,yang2026orthotryon}, various solutions have been proposed in the diffusion model domain for complex scene synthesis. To improve the results of combined generation, some methods \cite{kim2023dense,chen2024region,du2025textcrafter,bar2023multidiffusion,zhou2025dreamrenderer} focus on integrating instance-level control through multi-stage sampling, where the generation process is iteratively refined to ensure the correct placement and attributes of objects. Some methods \cite{xie2023boxdiff,chen2024training} have explored the use of a scoring function to guide the generation process, enabling zero-shot layout control that adapts to complex scene requirements. Another category of solutions \cite{guo2024initno,chefer2023attend} optimizes the initial noise to obtain better generation results. However, due to differences in image modeling approaches, these solutions are not suitable for VAR. In this paper, we introduce a training-free framework, SynVAR, tailored for the cross-scale pattern in the VAR paradigm. which establishes a collaborative correction mechanism of spatial and semantics, effectively enhancing the generation performance of VAR in complex scene synthesis.

\section{Method}
\subsection{Preliminaries}
VAR redefines autoregressive learning for images via next-scale prediction, inspired by human perception (coarse-to-fine hierarchy).  Instead of token-by-token generation, VAR generates multi-scale token maps ($r_0, r_1, ..., r_S$) autoregressively, starting from the coarsest resolution ($1 \times 1$) to the target resolution ($H \times W$). At each step $s$, the model predicts the entire token map $r_s$ conditioned on all previous scales $r_{0}, r_{1},...,r_{s-1}$:
\begin{equation}
    p\left(r_{0}, r_{1}, \ldots, r_{S}\right)=\prod_{s=1}^{S} p\left(r_{s} \mid r_{0}, r_{1}, \ldots, r_{s-1}\right)\cdot p(s_0)
\end{equation}

Using this sequence of residuals, the feature $f_{s}$ at the $s$-th scale step can be obtained from the following formula:
\begin{equation}
    f_{s}= f_{s-1} + \phi\left(r_{s-1},(h_{s}, w_{s})\right)
    \label{eq:f}
\end{equation}
where $\phi(\cdot)$ represents upsampling $r_{s-1}$ to resolution $\{h_s, w_s\}$. $f_{s} $ represents the cumulative sum of the upsampled sequence ($r_0, r_1, ..., r_{s-1}$). In the VAR generation process, once spatial or semantic errors occur in the early stage, this cross-scale dependency modeling can easily lead to the propagation and accumulation of errors, which is particularly common in complex scenarios.

\newcommand{\stx}[1]{\textcolor{red}{#1}}
\subsection{Overview}

\begin{figure*}[!h]
    % \vspace{-9mm}
    \centering
    \includegraphics[width=\textwidth]{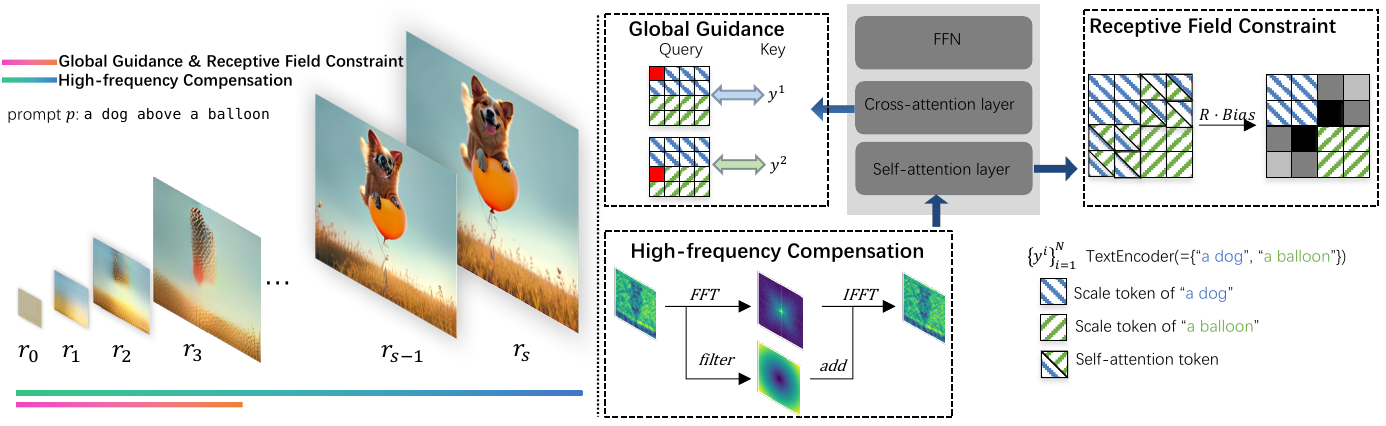}
    % \vspace{-6mm}
    \caption{Overall Framework of SynVAR. Global guidance and receptive field constraint are completed in the early steps of generation, while high-frequency compensation is continuously applied throughout the entire generation process. Specifically, global guidance introduces spatial priors into the cross-attention layers, receptive field constraint adjusts the perceptual scope of different regions in the self-attention layers, and high-frequency compensation enhances the high-frequency components of the features.}
    \label{Fig:method}
    % \vspace{-2mm}
\end{figure*}

As shown in Figure \ref{Fig:method}, we propose SynVAR, a spatial-semantics collaborative correction mechanism based on the generative characteristics of VAR. We use global guidance to explicitly introduce spatial priors to enhance the accuracy of object localization and recognition. Through receptive field constraints, we reduce interdependencies between different regions, preventing excessive information cross-over. We also adopt a high-frequency enhancement strategy to improve image details further. %Both global guidance and receptive field constraints are applied primarily during the early stage of generation, while high-frequency compensation spans the entire generation process to ensure that the final image is rich in details and appears natural.

\subsection{Global Guidance}

\begin{figure*}[t]
    % \vspace{-6mm}
    \centering
    \includegraphics[width=\textwidth]{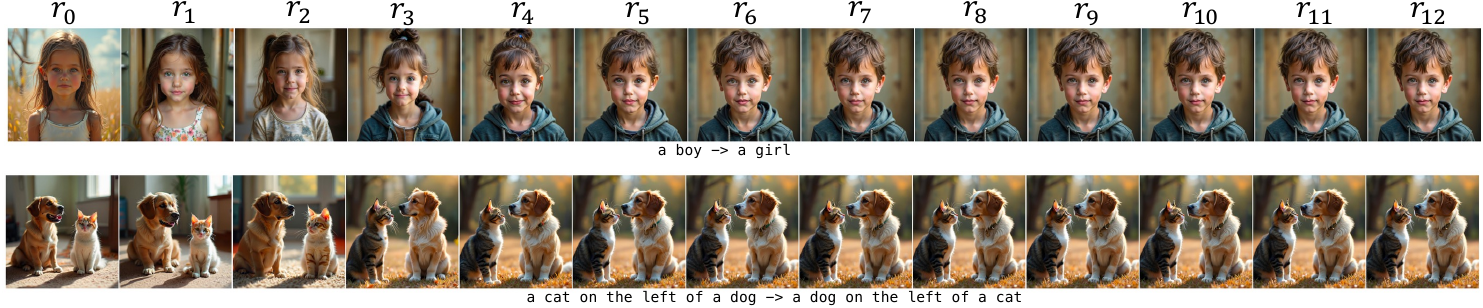}
    % \vspace{-6mm}
    \caption{Early fixation of semantic and spatial information. We force the replacement of semantic or spatial information in the VAR generation process, which can only affect the result in the early stage.}
    \label{Fig:Fix_Qualitative}
    % \vspace{-4mm}
\end{figure*}

\begin{wrapfigure}{r}{7cm}
    \centering
    \vspace{-8mm}
    \includegraphics[width=\linewidth]{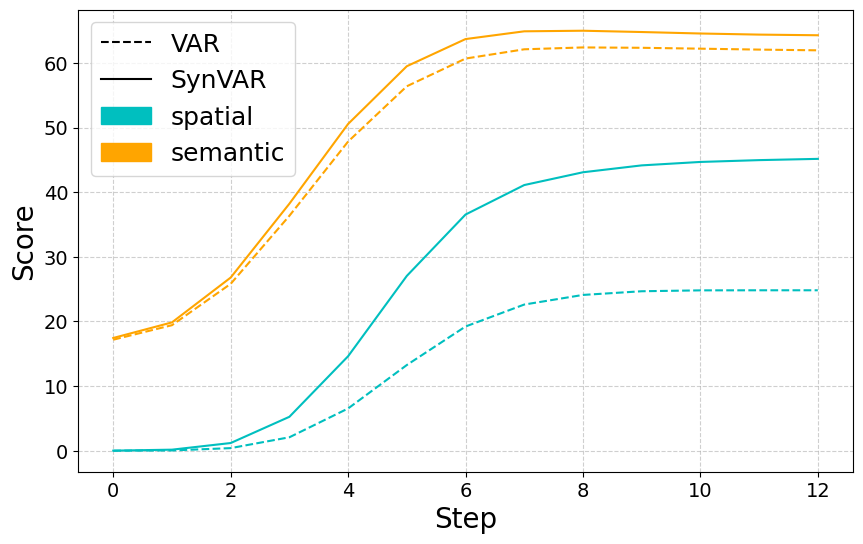}
    \vspace{-8mm}
    \caption{Quantitative metrics at different steps. On the Geneval benchmark, the semantic and spatial related indicators are close to convergence in the early stage.}
    \vspace{-8mm}
    \label{Fig:Fix_Quantitative}
\end{wrapfigure}

Some studies \cite{tian2024visual,guo2025fastvar,wang2025training} have shown that VAR primarily generates coarse-grained information in the early stages. As shown in Figure~\ref{Fig:Fix_Qualitative}, our further experiments reveal that there is an early fixation of semantic and spatial information during the VAR generation process. Specifically, when we artificially modify the spatial positions and semantic embeddings at various time steps during the generation, only changes made in the earlier stages have a meaningful impact on the final output. Quantitative analysis in Figure~\ref{Fig:Fix_Quantitative} indicates that spatial alignment and semantic consistency stabilize early in the process, with minimal adjustments taking place in the later stages. Therefore, to ensure structural integrity during the model's generation process and avoid generation errors, we apply a global guidance in the early stages of generation.

In global guidance, we break the original input prompt $p$ into a set of individual concepts $\{c^{i}\}_{i=1}^{N}$, where $N$ denotes the number of concepts in the prompt. We assign a global region $\{h^{i}, w^{i}\}_{i=1}^{N}$ for each concept and each individual concepts $\{c^{i}\}_{i=1}^{N}$ is encoded into a text embedding $\{y^{i}\}_{i=1}^{N}$ via text encoder, which is then projected into $K^{i}$ and $V^{i}$. Simultaneously, a query vector $Q^{i}$ is extracted the token map $r_{s-1}$. The detailed formulation is as follows:
\begin{equation}
    \begin{split}
        Q^{i}=\ell_{Q}\left(r_{s-1}\right), K^{i}=\ell_{K}(y^{i}) , V^{i}=&\ell_{V}(y^{i})  \\         
        r^{i}_{s-1}=\mathrm{Softmax}\left(\frac{Q^{i} K^{i T}}{\sqrt{d}}\right) V^{i},
    \end{split}
\end{equation}
where $\ell_{Q}$, $\ell_{K}$, $\ell_{V}$ are linear projections. Then, we crop and concatenate the token maps $r^{i}_{s-1}$ containing conceptual information according to their respective corresponding regions to achieve the injection of spatial information:
\begin{equation}
r^{i}_{s-1}(h^{i},w^{i})=\mathrm{Crop}\left(r^{i}_{s-1}, (h^{i},w^{i})\right)
\end{equation}
\begin{equation}
    r^{cat}_{s-1}= \mathrm{Concat}(r^{i}_{s-1}(h^{i},w^{i})_{i=1}^{N})
\end{equation}
% Finally, the semantic and spatial decision information brought by global guidance is used to obtain the residual prediction for the next step:
Finally, the sequence after the global guidance operation is used to obtain the output of the model:
% \begin{equation}
% r_{s}=\operatorname{interpolate}\left(r_{s-1}^{cat},\left(h_{s}, w_{s}\right)\right)+f_{s},
% \end{equation}
\begin{equation}
    f_{s}= f_{s-1} + \phi\left(r_{s-1}^{cat},(h_{s}, w_{s})\right)
\end{equation}

This guidance strategy significantly improves early space errors, reduce generation deviations, and enable the model to accurately render and correctly position multiple objects in complex scenes.

\subsection{Receptive Field Constraint}
\label{receptive field constraint}

\begin{wrapfigure}{r}{7cm}
    \centering
    \vspace{-8mm}
    \includegraphics[width=\linewidth]{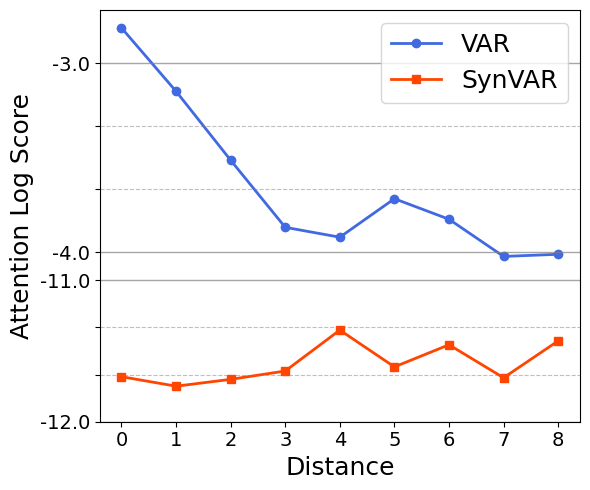}
    \vspace{-4mm}
    \caption{Attention scores between different targets.  the level of dependency among them remains consistently high, particularly in adjacent regions.}
    \vspace{-6mm}
    \label{Fig:regional_coupling}
\end{wrapfigure}

The VAR model enhances the receptive field range effectively by using two-dimensional sequences, which allows for more effective global contextual information. However, as illustrated in Figure \ref{Fig:regional_coupling}, we observe that when multiple target regions are present, the attention dependencies between different regions remain high, especially when they are close to each other. This leads to a high coupling of information between different object regions, resulting in semantic confusion. To address this issue, we propose a receptive field constraint mechanism to regulate the attention dependencies between tokens, helping VAR obtain the correct semantic information.

In receptive field constraint, we introduce a gaussian mask into the self-attention to calculate the relative spatial proximity of tokens in the early stage of generation. 
Specifically, a bias term is computed based on their Euclidean distance in the feature map, the formula is as follows:
\begin{equation}
    Bias[q,k] = \exp\left(-\frac{||p_q - q_k||_2^2}{\sigma^2}\right)
    \label{eq:bias}
\end{equation}
where $p_q$ and $q_k$ represent the coordination of the corresponding token of the query and key in the feature map. $\sigma$ is a predefined decay factor controlling the spatial sensitivity. Then we define a binary mask to represent the region relationship of the query and key:
\begin{equation}
    R[q, k]=\mathbb{I}\left(\forall m, q \notin \mathcal{I}_{m} \vee k \notin \mathcal{I}_{m}\right)
    \label{eq:r}
\end{equation}
where $\mathcal{I}_m$ represents the set of token indices belonging to the $m$-th region.
Lastly, we get the gaussian mask and use it for the modulation of the attention map:
% \begin{equation}
%     M=R\cdot Bias
% \end{equation}
\begin{equation}
    A^{\prime}=\mathrm{Softmax}(\frac{QK^\top+(R\cdot Bias)}{\sqrt{d}})
    \label{eq:self-attention}
\end{equation}

The advantage of this method is that it can retain the necessary context information while performing semantic decoupling. The greater the distance, the greater the attenuation, thus avoiding excessive cross-region interactions. We also tried a variety of different methods to implement the receptive field constraint, see Supplementary Material for details.

\subsection{High-frequency Compensation}

\begin{wrapfigure}{r}{7cm}
    \centering
    \vspace{-8mm}
    \includegraphics[width=\linewidth]{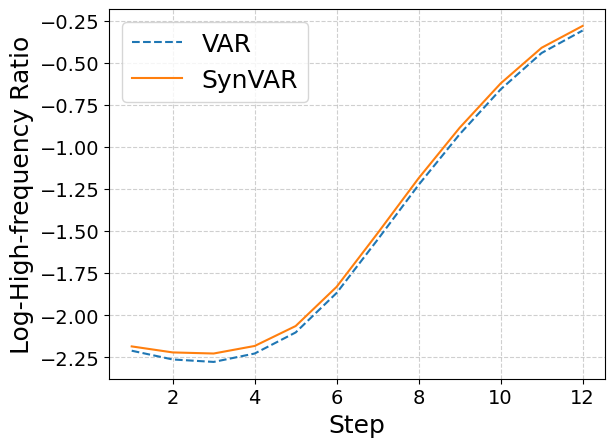}
    \vspace{-4mm}
    \caption{High-frequency ratio. As generation progresses, high-frequency components increase. A small intensity coefficient is used due to its sensitivity.}
    \vspace{-8mm}
    \label{Fig:frequency_ratio}
\end{wrapfigure}

We decompose the feature map into low-frequency and high-frequency components to better observe the evolution of image details during the generation process. In the early stages, the model primarily generates low-frequency components, which capture the broad structure and general shape of the image. As the generation process progresses, the proportion of high-frequency information gradually increases, focusing on finer details and textures. This behavior is illustrated in Figure \ref{Fig:frequency_ratio}, where the model begins with low-frequency information and progressively shifts towards high-frequency optimization, enhancing the image’s texture and details. To this end, we propose controlling the intensity of the high-frequency components in the feature maps during the generation process, thereby obtaining high-fidelity images with more richly detailed content.

We begin by defining a high-pass filter with a linear gain controlled by the enhancing ratio $\delta$, which allows us to adjust the strength of the high-frequency components:
\begin{equation}
    D(u, v) = \sqrt{(u - u_0)^2 + (v - v_0)^2}
\end{equation}
% \vspace{-6mm}
\begin{equation}
    HF(u, v) = 1 + \delta \cdot D(u, v)
\end{equation}
where $(u_0, v_0)$ denotes the center of the frequency spectrum, typically $u_0 = W/2$ and $v_0 = H/2$ for an image of size $H \times W$.
We then perform a Fourier transform on the input of each transformer layer to convert it to the frequency domain and enhance its high-frequency components through the above high-pass filter. Finally, we perform an inverse Fourier transform on the filtered signal to restore it back to the spatial domain:
\begin{equation}
    \tilde{\mathbf{F}} = \mathcal{F}^{-1} \left[ \mathcal{F}[\mathbf{F}] \cdot HF \right]
\end{equation}
where $\mathbf{F} \in \mathbb{R}^{H \times W \times C}$ denotes the input feature map.
$\mathcal{F}[\cdot]$ and $\mathcal{F}^{-1}[\cdot]$ denote the 2D Fourier and inverse Fourier transform, respectively.
$HF \in \mathbb{R}^{H \times W}$ represents a high-pass filter mask in the frequency domain.
$\tilde{\mathbf{F}}$ is the filtered feature map after enhancing high-frequency components.
In this way, we are able to significantly enhance the details and texture information of the image, ensuring the richness and fineness of the details in the later stages of generation.
% \vspace{-4mm}

The pseudocode for SynVAR is shown in Algorithm \ref{alg:synvar_simplified_overview}.

\newcommand{\upsamplephi}{\phi}
% Define mathcal F for Fourier Transform
\newcommand{\fourierF}{\mathcal{F}}

\newcommand{\Requirex}[2]{%
  \item[\textbf{#1}] #2
}
\begin{algorithm}[H] % Use [H] for "here" placement, or [htbp]
\caption{SynVAR Algorithm Overview}
\label{alg:synvar_simplified_overview}
\begin{algorithmic}[1]
\Requirex{Input:}{input prompt $p$, global regions $\{h^{i}, w^{i}\}_{i=1}^{N}$}, individual concepts $\{c^{i}\}_{i=1}^{N}$, Transformer block input feature $\textbf{F}$.

\Requirex{Hyperparameters:}  {selected steps $S_{steps}$, resolution of scales $\{h_s, w_s\}_{i=1}^{S}$}, decay factor $\sigma$, enhancing ratio $\delta$.

\Requirex{Model:}{linear projections $\ell_Q,\ell_K,\ell_V$ }, Text Encoder TE($\cdot$). $\mathcal{F}[\cdot]$ and $\mathcal{F}^{-1}[\cdot]$ is the 2D Fourier and inverse Fourier transform.

\State $\{y^{i}\}_{i=1}^{N} \leftarrow TE(\{c^{i}\}_{i=1}^{N})$
\For{$s = 1, \cdots,  S$} 

\State $    HF \leftarrow 1 + \delta \cdot \sqrt{(u - u_0)^2 + (v - v_0)^2} $ \Comment{\textbf{High-frequency Compensation}}

\State $    \tilde{\mathbf{F}} = \mathcal{F}^{-1} \left[ \mathcal{F}[\mathbf{F}] \cdot HF \right]$

\State $r_{s-1}\leftarrow \tilde{\mathbf{F}}$
    
    \If{$s \in S_{steps}$}
        % \For{$m = 1, \cdots, N$}  
            \State $Bias \leftarrow \exp\left(-\frac{||p_q - q_k||_2^2}{\sigma^2}\right)$ \Comment{\textbf{Receptive Field Constraint} in self-attention}
            \State $R \leftarrow \mathbb{I}\left(\forall m, q \notin \mathcal{I}_{m} \vee k \notin \mathcal{I}_{m}\right)$
        % \EndFor
        \State $r_{s-1} \leftarrow \mathrm{Softmax}(\frac{QK^\top+(R\cdot Bias)}{\sqrt{d}})V
    \label{eq:self-attention}$
    
        \For{$i = 1, \cdots, N$}  \Comment{\textbf{Global Guidance} in cross-attentions}
            \State $Q^i \leftarrow \ell_Q(r_{s-1})$; $K^i \leftarrow \ell_K(y^i)$; $V^i \leftarrow \ell_V(y^i)$
            \State $r^i_{s-1} \leftarrow \mathrm{Softmax}\left(\frac{Q^i K^{iT}}{\sqrt{d}}\right) V^i$ 
        \EndFor
       \State $r^{cat}_{s-1} \leftarrow \mathrm{Crop}(r^i_{s-1}(h^i, w^i))$ 
       \State $f_{s} \leftarrow f_{s-1} + \phi\left(r_{s-1}^{cat},(h_{s}, w_{s})\right)$
       \Else 
        \State $f_{s} \leftarrow f_{s-1} + \phi\left(r_{s-1},(h_{s}, w_{s})\right)$
    \EndIf
% \State update $f_s$ using Eq \ref{eq:f}
% \If{$s<S$}
% \State $\mathbf{F}\leftarrow \text{downsample}(f_s,(h_{s+1},w_{s+1})) $  %表示成将f_s下采样
% \EndIf
\EndFor 
\Return Decoder($f_s$)
\end{algorithmic}
\end{algorithm}

% \vspace{-2mm}
\section{Experiments}

\subsection{Experiment Setting}
\noindent{\textbf{Implementation Details.}} For the influence step of global guidance and receptive field constraints, we choose to operate at step=2. The decay coefficient $\sigma$ of the receptive field constraint is set to 0.5. The intensity control coefficient $\delta$ for high-frequency compensation is set to 0.01. Other parameters are consistent with the compatible base model. For large-scale quantitative evaluations, we leverage MLLM for automated global region and concept division, see Appendix for usage (Notably, our framework is MLLM-free. MLLMs are only employed for large-scale evaluation and user convenience, with additional experiments detailed in the Appendix). All experiments are implemented on a single A6000. We also provide runnable code in the Appendix for understanding and reproduction. 

\newcommand{\pub}[1]{\textup{/(#1)}}          
\newcommand{\textgr}[1]{\textcolor[HTML]{008B45}{#1}}
\begin{table*}[!t]
	\centering
	\resizebox{\textwidth}{!}{
	\setlength{\tabcolsep}{3.8pt}
	\footnotesize
	\renewcommand\arraystretch{1.1}
	\begin{tabular}{cccccccccccc} % 12列（1方法 + 5Geneval + 6T2ICompBench）
		\toprule[1pt] % 顶部粗线
		\rowcolor[HTML]{FAFAFA}
		& \multicolumn{5}{c}{Geneval} 
		& \multicolumn{6}{c}{T2I-CompBench} \\
		\cmidrule(lr){1-1} \cmidrule(lr){2-6} \cmidrule(lr){7-12}
		Method 
            & $Two\_object\, \uparrow$
		& $Counting\, \uparrow$ 
		& $Position\, \uparrow$ 
		& $Attribute\_Binding\, \uparrow$ 
		& $Overall\, \uparrow$
		& $Color\, \uparrow$ 
		& $Shape\, \uparrow$ 
		& $Texture\, \uparrow$ 
		& $Spatial\, \uparrow$ 
            & $Complex\, \uparrow$
		& $Overall\, \uparrow$ \\ 
		\midrule[0.8pt] % 中间细线
		
		\rowcolor[HTML]{F8FFF8} Infinity \hspace{0.2em} \scriptsize{\textcolor{gray}{[CVPR2025]}}
		& 79.80 & 58.13 & 26.00 & 58.00 & 55.48 & 74.99 & 48.59 & 62.90 & 24.13 & 38.89 & 49.90 \\
        \rowcolor[HTML]{F8FFF8} +Densediffusion 
		& 75.00 & 19.69 & 21.75 & 52.50 & 42.23 & 72.42 & 48.39 & 62.64 & 24.78 & 38.55 & 49.36 \\
         \rowcolor[HTML]{F8FFF8} +RAG-Diffusion 
		& 70.71 & 40.31 & 30.00 & 44.75 & 46.44 & 59.86 & 34.85 & 56.11 & 27.05 & 33.06 & 42.19 \\
		\rowcolor[HTML]{F8FFF8} +Ours 
		& 91.92 & 58.25 & 63.00 & 70.00 & 70.79 & 75.21 & 51.85 & 66.23 & 46.00 & 39.02 & 55.66 \\
		\rowcolor[HTML]{F8FFF8} \multicolumn{1}{c}{Ours vs Infinity} 
		& \textgr{+12.12} & \textgr{+0.12} & \textgr{+37.00} & \textgr{+12.00} 
		& \textgr{+15.31} & \textgr{+0.22} & \textgr{+3.26} & \textgr{+3.33} 
		& \textgr{+21.87} & \textgr{+0.13} & \textgr{+5.76} \\
		
		\midrule[0.3pt] % 模块分隔线
		
		\rowcolor[HTML]{F8FBFF} Switti \hspace{0.2em} \scriptsize{\textcolor{gray}{[CVPR2025]}}
		& 73.99 & 48.12 & 14.25 & 28.00 & 41.09 & 74.83  & 52.00 & 64.42 & 19.09 & 37.72 & 49.61 \\
         \rowcolor[HTML]{F8FBFF} +Densediffusion 
		& 65.91 & 18.44 & 19.00 & 27.75 & 32.77 & 71.25 & 49.60 & 61.15 & 17.49 & 36.68 & 47.23 \\
         \rowcolor[HTML]{F8FBFF} +RAG-Diffusoin 
		& 36.62 & 3.44 & 21.00 & 18.00 & 19.76 & 55.92 & 37.81 & 53.65 & 12.42 & 31.33 & 38.22 \\
		\rowcolor[HTML]{F8FBFF} +Ours 
		& 75.25 & 49.06 & 16.25 & 40.50 & 45.27 & 78.54 & 54.67 & 68.10 & 22.14 & 38.77 & 52.44 \\
		\rowcolor[HTML]{F8FBFF} \multicolumn{1}{c}{Ours vs Switti} 
		& \textgr{+1.26} & \textgr{+0.94} & \textgr{+2.00} & \textgr{+12.50} 
		& \textgr{+4.18} & \textgr{+3.71} & \textgr{+2.67} & \textgr{+3.68} 
		& \textgr{+3.05} & \textgr{+1.05} & \textgr{+2.83} \\
		
		\bottomrule[1pt] % 底部粗线
	\end{tabular}
	}
    % \vspace{-2mm}
        \vspace{2mm}
        \caption{Quantitative results on Geneval and T2I-CompBench benchmark.}
        \label{Tab:Quantitative Comparison}
\end{table*}

\begin{figure*}[!t]
    % \vspace{-6mm}
    \centering
    \includegraphics[width=\textwidth]{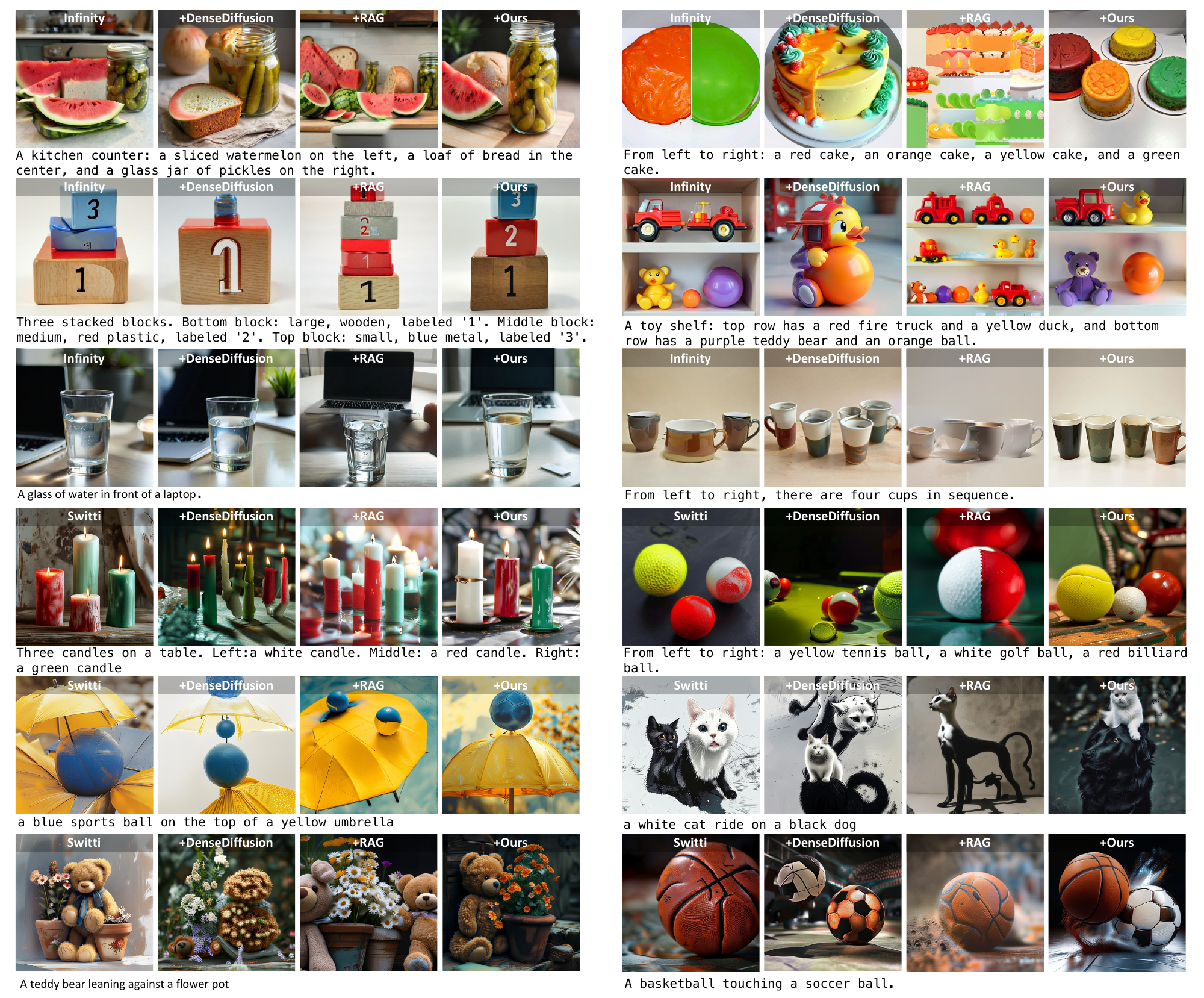}
    % \vspace{-6mm}
    \caption{Qualitative comparison of different methods for adapting VAR-based text-to-image models.}
    \label{Fig:Qualitative_comparison}
    % \vspace{-6mm}
\end{figure*}

\noindent{\textbf{Baselines and Benchmark.}} 
Since SynVAR is designed as a plug-and-play solution, we integrate it with existing VAR-base methods infinity and Switti for experiments. We apply consistent parameter settings across all models to ensure fair comparison. For the generation quality, we conduct experimental evaluation on two widely used benchmark Geneval \cite{ghosh2023geneval} and T2I-CompBench \cite{huang2023t2i}.

\subsection{Main Results}

\noindent{\textbf{Quantitative Comparison.}} As shown in Table \ref{Tab:Quantitative Comparison}, existing enhancement methods generally have a negative impact on the performance of the original model, suggesting their limited effectiveness in addressing the inherent weaknesses of VAR-based methods under complex compositional prompts. In contrast, integrating SynVAR leads to consistent and significant improvements. On the Geneval dataset, SynVAR boosts the overall performance of the Infinity model by 27.6\% and that of Switti by 10.2\%. Notably, SynVAR achieves clear gains on both spatial metrics (e.g., Position in Geneval and Spatial in T2I-CompBench) and semantic metrics (e.g., Attribute\_Binding in Geneval and Shape/Texture in T2I-CompBench), demonstrating its capability to simultaneously reduce spatial errors and semantic mismatches.
%\textcolor{blue}{some specific aspects like the significant color improvement should be further considered. Why some aspects can be improved more than others? Do they relate to the motivation of the methods?}

\noindent{\textbf{Qualitative Comparison.}} As illustrated in Figure \ref{Fig:Qualitative_comparison}, existing approaches fail to resolve spatial dislocation or semantic confusion when generating images from complex prompts. For instance, objects often appear in incorrect positions or exhibit inconsistent appearances. By contrast, SynVAR achieves more coherent spatial arrangements and preserves semantic fidelity more effectively. It not only reduces spatial errors and semantic confusion but also produces visually pleasing results with sharper structures and finer textures. Moreover, the generated global scene appears more harmonious and contextually aligned with the input prompt, indicating that SynVAR successfully captures both local detail and holistic scene layout. In addition, we have demonstrated more practical and interesting usages in the Appendix, such as overlapping, multi-objective fine-grained interaction, and the utilization of irregular masks.

\begin{wrapfigure}{r}{7cm}
    \centering
    \vspace{-8.5mm}
    % \hspace{-60mm}
    \includegraphics[width=\linewidth]{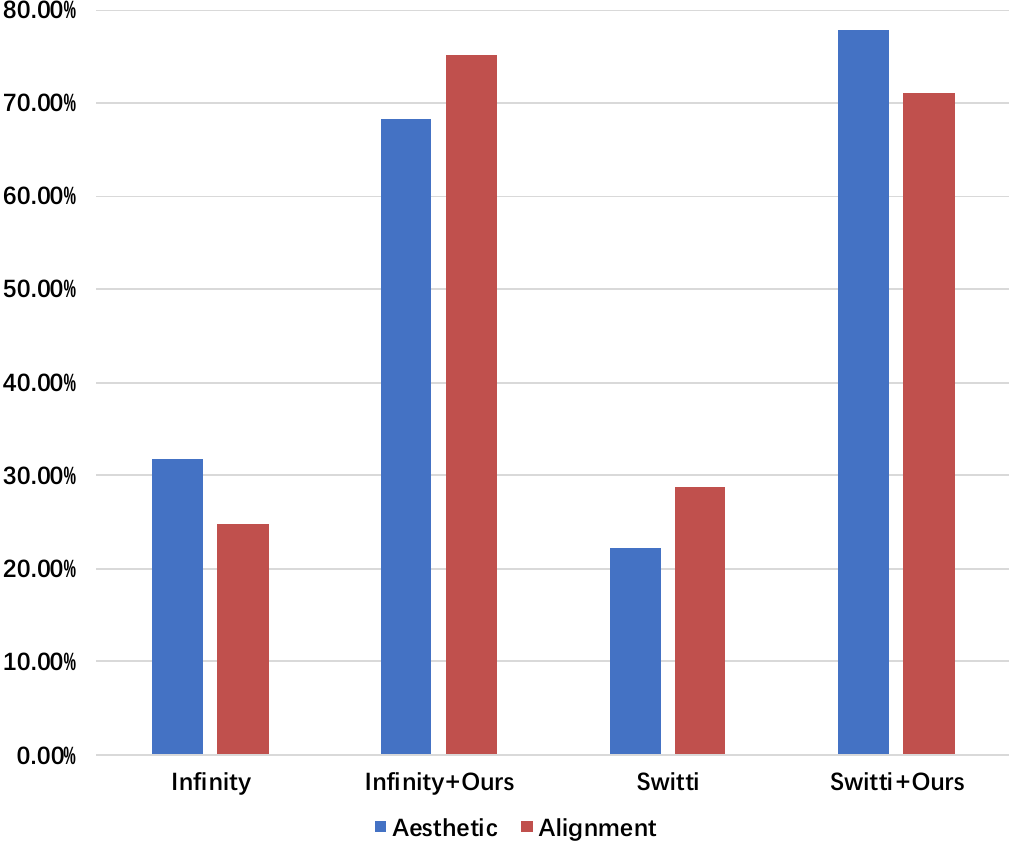}
    \vspace{-3mm}
    \caption{User study on aesthetics and text-image alignment. Our approach shows a significant improvement.}
    \label{Fig:User_study}
    \vspace{-8mm}
\end{wrapfigure}

\noindent{\textbf{User Study.}}
To further validate the effectiveness of SynVAR from a human perspective, we conducted a user study. We randomly selected 20 prompts from the Geneval and T2Icombench to assess the aesthetic quality and text-image alignment (including both semantic and spatial position) of the generated images. During the evaluation, we randomly displayed an image pair and its corresponding text description, and users selected the image that best matched the prompt based on their judgment. The results in Figure \ref{Fig:User_study} show that more than 68\% of users chose the image enhanced by SynVAR in terms of both aesthetics and alignment, indicating that SynVAR not only improves objective performance metrics but also aligns more closely with human visual perception and interpretive judgment.

\subsection{Ablation Study}

\begin{table}[h]
    \centering
    \resizebox{\textwidth}{!}{
        \begin{tabular}{c c c c c c}
        \toprule
        Methods & Two\_object & Counting & Position & Attribute\_Binding & Overall \\
        \midrule
        w/o Global Guidance & 74.49 & 66.25 & 23.75 & 49.00 & 53.37 \\
        w/o Receptive Field Constraint & 80.30 & 21.25 & 36.25 & 58.25 & 49.01 \\
        w/o High-Frequency Compensation & 86.38 & 55.81 & 59.50 & 62.75 & 66.11 \\
        \midrule
        SynVAR & \textbf{91.92} & \textbf{58.25} & \textbf{63.00} & \textbf{70.00} & \textbf{70.79} \\ 
        \bottomrule
        \end{tabular}
    }
     \vspace{2mm}
     \caption{Ablation performance of SynVAR on the Geneval.}
     \label{Tab:Ablation_Study}
\end{table}

\begin{figure*}[h]
    % \vspace{-8mm}
    \centering
    \includegraphics[width=\textwidth]{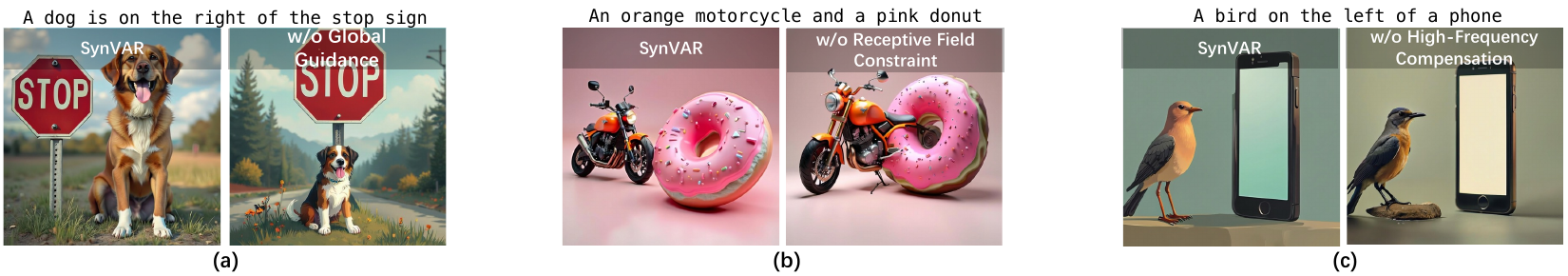}
    % % \vspace{-6mm}
    \caption{Qualitative analysis of global guidance, receptive field constraint and high-frequency compensation.}
    \label{Fig:module_ablation}
    % \vspace{-3mm}
\end{figure*}

\noindent{\textbf{Effectiveness of Global Guidance.}}
As shown in Table \ref{Tab:Ablation_Study},removing the global guidance mechanism reduces the spatial position accuracy by 62.3\%, and also causes a decrease in other indicators. This means that global guidance effectively improves the correctness of the spatial structure by explicitly injecting spatial priors, laying a solid foundation for the subsequent generation process, thereby improving the image indicators in all aspects. Figure \ref{Fig:module_ablation}(a) shows the visual effect of global guidance in ensuring the correct spatial structure.

\noindent{\textbf{Effectiveness of Receptive Field Constraint.}}
The receptive field constraint helps reduce excessive interactions between regions, preserving object generation independence. Table \ref{Tab:Ablation_Study} shows that it significantly lowers interference, allowing each region to generate its object independently, improving the counting metric. Figure \ref{Fig:module_ablation}(b) visually demonstrates this effect, where regions decouple their target objects while preserving semantic integrity.

\begin{figure}[h]
    \centering
    \includegraphics[width=\linewidth]{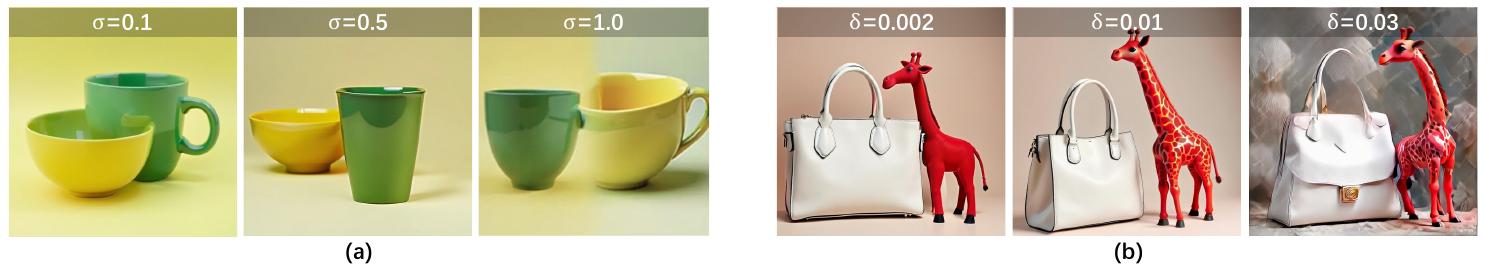}
    % \vspace{-6mm}
    \caption{Qualitative analysis of decay coefficient $\sigma$ and  intensity control coefficient $\delta$.}
    \label{Fig:parameter_ablation}
    % \vspace{-6mm}
\end{figure}

An ablation study on the decay coefficient $\sigma$ in the receptive field constraint reveals a non-linear relationship with performance. From the data in Table \ref{Tab:Ablation sigma}, as $\sigma$ increases, performance improves initially but declines beyond a point. This suggests that excessive constraint hinders performance, while a moderate $\sigma$ balances region independence and necessary information transfer, enhancing output quality. This is shown in Figure \ref{Fig:parameter_ablation}(a), where a balanced decay coefficient produces better results.

\noindent
\begin{minipage}{\textwidth}
\vspace{6mm}
\begin{minipage}[t]{0.5\textwidth}
\makeatletter\def\@captype{table}
\resizebox{1.0\linewidth}{!}{
        \begin{tabular}{c|ccccc}
        \toprule
        Methods & Two\_object & Counting & Position & Attribute\_Binding & Overall \\
        \midrule
        $\sigma=0.1$ & 80.56 & 46.25 & 34.50 &57.75 & 54.76 \\
        % \midrule
        $\sigma=0.3$ &88.38 & 54.69 & 53.25 & 65.50 & 65.46\\
        % \midrule
        $\sigma=0.5$ & \textbf{91.92} & \textbf{58.25} & \textbf{63.00} & \textbf{70.00} & \textbf{70.79}\\
        % \midrule
        $\sigma=0.7$ & 90.66 & 57.50 & 60.25 & 67.50 & 68.98 \\
        % \midrule
        $\sigma=1.0$ & 90.66 & 55.00 & 57.25 & 67.00 & 67.48 \\
        \bottomrule
        \end{tabular}
    }
% \vspace{-3mm}
\caption{Ablation experiment of decay coefficient $\sigma$ in receptive field constraint.}
\label{Tab:Ablation sigma}
\end{minipage}
\begin{minipage}[t]{0.5\textwidth}
\makeatletter\def\@captype{table}
\resizebox{1.0\linewidth}{!}{
    \begin{tabular}{c|ccccc}
    \toprule
    Methods & Two\_object & Counting & Position & Attribute\_Binding & Overall \\
    \midrule
    $\delta=0.002$ & 90.40 & 56.56 & 57.00 &65.00 & 67.24 \\
    $\delta=0.005$ & 89.14 & 55.00 & 60.50 &69.25 & 68.47 \\
    % \midrule
    $\delta=0.01$ & 91.92 & \textbf{58.25} & 63.00 & \textbf{70.00} & \textbf{70.79}\\
    % \midrule
    $\delta=0.02$ & \textbf{92.17} & 54.37 & \textbf{65.50} & 67.00 & 69.76\\
    % \midrule
    $\delta=0.03$ & 90.66 & 51.25 & 63.75 & 66.50 & 68.04\\
    \bottomrule
    \end{tabular}
}
% \vspace{-3mm}
\caption{Ablation experiment of intensity control coefficient $\delta$ in High-frequency Compensation.}
\label{Tab:Ablation delta}
\end{minipage}
% \vspace{-2mm}
% \vspace{4mm}
\end{minipage}

\noindent{\textbf{Effectiveness of High-frequency Compensation.}}
High-frequency compensation enhances fine-grained details like textures and intricate structures. While its impact on overall metrics is modest, it noticeably improves image quality. Figure \ref{Fig:module_ablation}(c) highlights its effect, refining details, especially in complex areas. We also conducted a user study to evaluate our advantages in terms of image quality (e.g., Aesthetics), see the supp for details.

An analysis of the intensity control coefficient $\delta$ reveals that while performance remains stable with minor fluctuations, excessive enhancement can introduce artifacts, as shown in Table \ref{Tab:Ablation delta} and Figure \ref{Fig:parameter_ablation}(b). We chose $\delta$=0.01 as an optimal value, balancing detail enhancement with image integrity.

\begin{table}[h]
    \centering
    \resizebox{\textwidth}{!}{
        \begin{tabular}{c c c c c c}
        \toprule
        Step & Two\_object & Counting & Position & Attribute\_Binding & Overall \\
        \midrule
        $1$ & 83.33 & 57.50 & 45.00 & 66.00 & 63.43 \\
        $2$ & 91.92 & \textbf{58.25} & 63.00 & \textbf{70.00} & \textbf{70.79} \\
        $3$ & 81.06 & 52.19 & 36.25 & 61.25 & 57.69 \\
        $1,2$ & 89.14 & 54.06 & 68.25 & 68.00 & 69.86 \\
        $1,3$ & 86.11 & 50.94 & 55.00 & 67.50 & 64.89 \\
        $1,2,3$ & 92.68 & 47.81 & 74.50 & 63.25 & 69.56 \\
        $1,2,3,4$ & \textbf{93.69} & 39.38 & 75.00 & 60.00 & 67.02 \\
        $1,2,3,4,5$ & 91.41 & 32.81 & \textbf{75.25} & 58.50 & 64.49 \\
        \bottomrule
        \end{tabular}
    }
    \vspace{2mm}
    \caption{Quantitative comparison of global guidance and receptive field constraints at different steps.}
    \label{Fig:step_ablation}
\end{table}

\begin{figure*}[!t]
    % \vspace{-12mm}
    \centering
    \includegraphics[width=\textwidth]{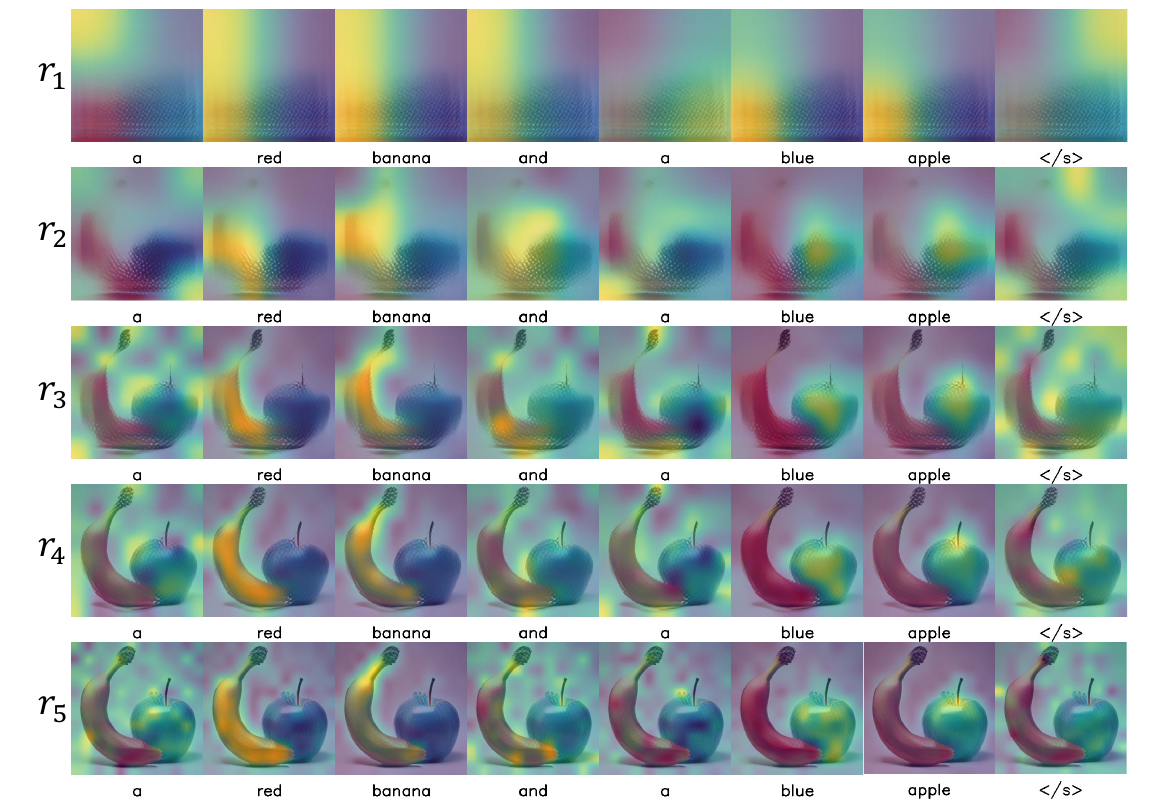}
    % \vspace{-6mm}
    \caption{Visual analysis of the attention area during the generation process.}
    % \vspace{-6mm}
    \label{Fig:Visual_attention}
\end{figure*}

\noindent{\textbf{Selection of Early Steps for Global Guidance and Receptive Field Constraints.}} 
We conduct ablation studies to verify the optimal stage for applying global guidance and receptive field constraints during the generation sequence. As shown in Table \ref{Fig:step_ablation}, extending these operations to multiple iterations leads to performance degradation, but the overall trend remained stable. 

We also visualize the attention regions during the VAR generation process, as illustrate in Figure \ref{Fig:Visual_attention}. It can be observed that the attention focus stabilizes at very low resolutions, with minimal changes in subsequent steps. This supports the findings in Table \ref{Fig:step_ablation}, indicating that injecting structured priors at the critical decision window (step = 2) enables accurate correction of error sources while maintaining the coherence of the VAR generation.

\noindent{\textbf{Inference Time Cost.}}
As shown in Table \ref{tab:time_cost}, we tested the single-image inference time consumption of different components. It can be seen that the inference overhead of Global and Receptive Field Constraints is almost negligible compared to Vanilla VAR. High-Frequency Compensation, due to the frequency domain conversion involved, incurs additional time overhead, but this is traded off for improved aesthetic quality. Since our method does not involve any structural changes, SynVAR can seamlessly integrate with existing 
\begin{wraptable}{r}{8cm}
    % \vspace{-6mm}
    \centering
        \resizebox{\linewidth}{!}{
        \begin{tabular}{c|c}
        \toprule
            Methods & Inference Time Cost (seconds/image)\\
            \midrule
            Vanilla VAR & 2.60 \\
            % \midrule
            w/ Global Guidance & 2.61 \\
            w/ Receptive Field Constraint & 2.61 \\
            w/ High-Frequency Compensation & 3.08 \\\
            
            SynVAR & 3.10 \\
            SynVAR + Fastvar & 2.27 \\
            \bottomrule
        \end{tabular}
        }
    % \vspace{-4mm}
    \caption{Single-image inference time consumption of different SynVAR components.}
    \label{tab:time_cost}
    \vspace{-6mm}
\end{wraptable}
VAR acceleration schemes (such as Fastvar~\cite{guo2025fastvar}), thereby ensuring image quality improvements while avoiding increasing inference overhead. We have also provided performance metrics compatible with Fastvar in the Appendix.

\section{Conclusion}
\label{sec:Conclusion}
In this paper, we conduct an in-depth analysis of the failure mechanisms of VAR in complex scene generation and propose the first training-free enhancement framework for VAR, dubbed SynVAR. SynVAR implements a collaborative correction mechanism for spatial and semantic control through three key components: global guidance, receptive field constraints, and high-frequency compensation. Global guidance and receptive field constraints work during the early stages of generation, utilizing spatial priors and gaussian masks to capture precise spatial and semantic information, effectively preventing the propagation and accumulation of errors. High-frequency compensation enhances high-frequency features, further improving the detail and generating high-fidelity images. Extensive experimental results demonstrate that SynVAR seamlessly integrates with various VAR-based text-to-image models and significantly improves generation performance. In the future, we will try to extend SynVAR to more types of generative models, hoping to provide more valuable insights to the community.
    
\section{Limitation and Future Work}
Our framework mainly has two limitations. First, the capability of our framework to better align the semantic and spatial relations is limited by the prior of the VAR foundation model. If the semantic concepts are missed in the prior or the spatial relation deviation is unreasonable, our correction mechanism is deficient at such imprinted limitations of the foundation model. Second, for the best performance, the optimal early step of global guidance and receptive field constraints through experiments and empirical settings. In the future, we expect to develop a lightweight fine-tuning and auto-selection strategies to edit the model prior and automatically select the optimal step for each single image.
\section*{Acknowledgment}
This work was supported by Natural Science Foundation of Jiangsu Province: BK20241198, the Gusu Innovation and Entrepreneur Leading Talents: No. ZXL20 24362 and Natural Science Foundation of China: No. 62406135, Nanjing University-China Mobile Communications Group Co. Ltd. Joint Institute.

\bibliographystyle{splncs04}
\bibliography{main}
% \clearpage
% \setcounter{page}{1}
% \maketitlesupplementary

%更多的方法细节：1.伪代码 2.其他的方法细节(待定)
% \clearpage
% \appendix

\section{More Model Details}

\subsection{Explanation of Global Region Division}
\begin{figure}[h]
    \vspace{-8mm}
    \centering
    \includegraphics[width=0.8\textwidth]{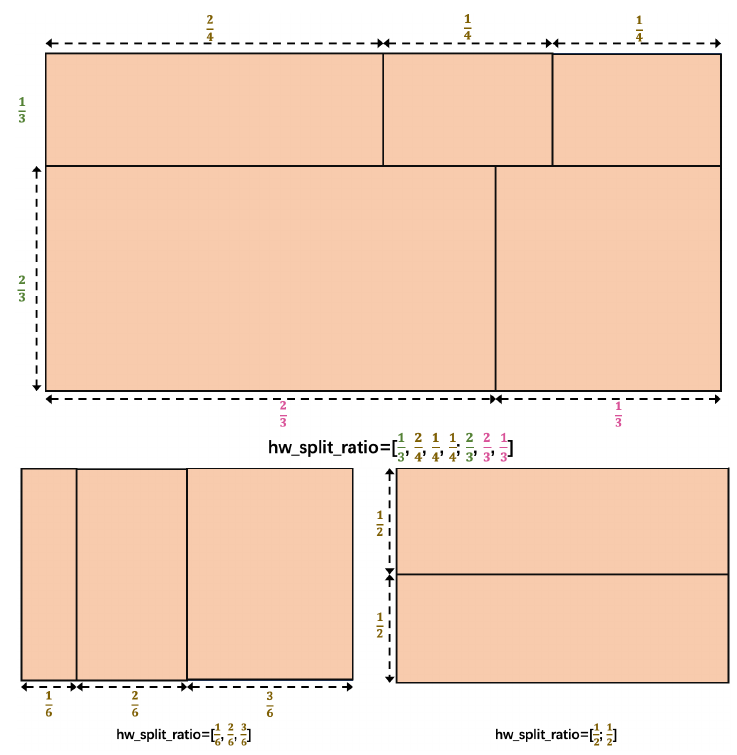}
    \vspace{-4mm}
    \caption{Several schematic diagrams of regional division.}
    \label{Fig:region_example}
     \vspace{-6mm}
\end{figure}

In Figure \ref{Fig:region_example}, we present an example image of the global region division $\{h^{i}, w^{i}\}_{i=1}^{N}$. The global region division ensures coverage of the entire area, ensuring that the generated content conforms to the specified spatial layout while ensuring interaction and integration of adjacent areas. This collaborative design significantly improves the attribute fidelity of the generated images while promoting harmonious coordination across different regions, resulting in more cohesive and visually consistent outputs.

\subsection{Automatic Global Regional and Concept Division}
\begin{figure*}[!ht]
    \vspace{-6mm}
  \centering
  \includegraphics[scale=0.35]{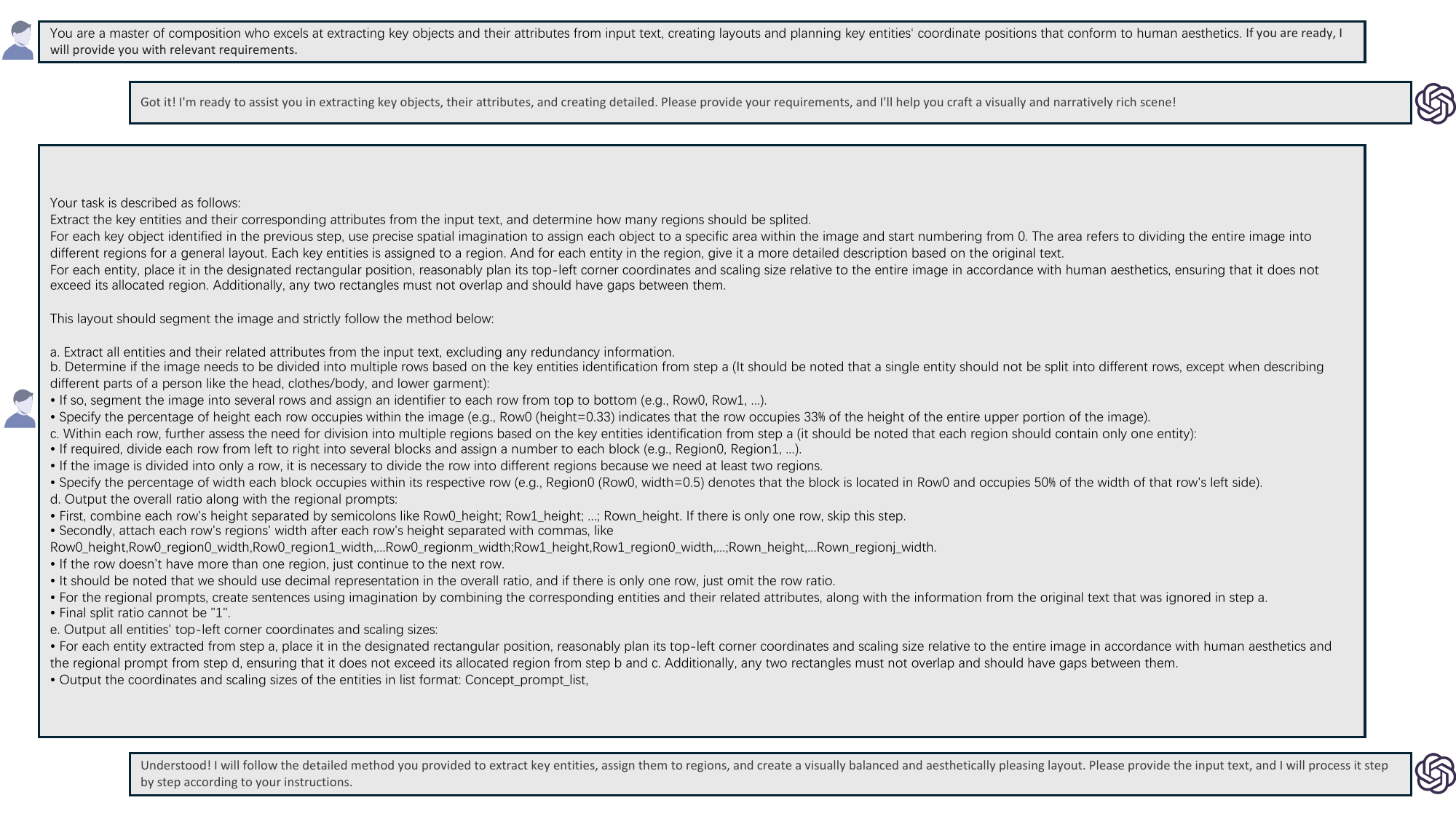}
  \hspace{1in}
  \includegraphics[scale=0.35]{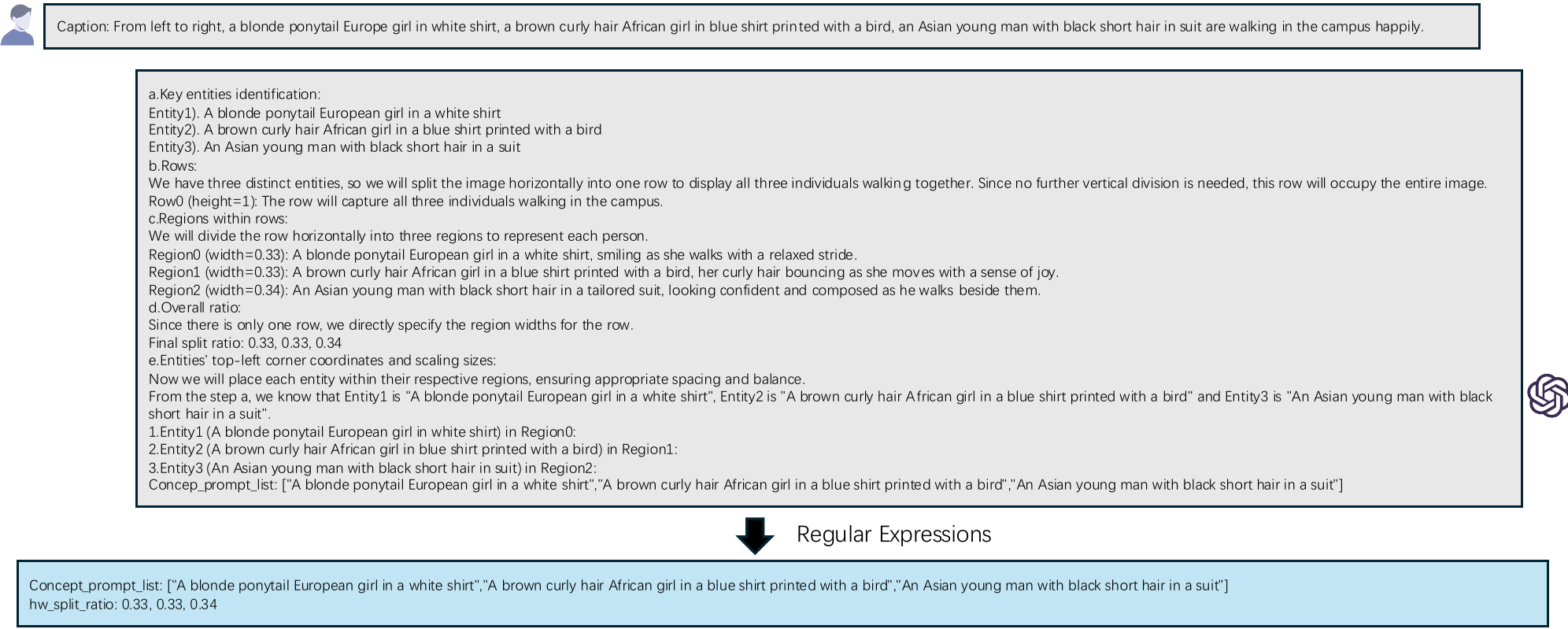}
  \caption{Automatic Global Region and Concept Division.}
  \label{Fig:MLLM}
  \vspace{-6mm}
\end{figure*}

In order to facilitate quantitative testing and user use, we use MLLM to automatically segment spatial regions and generate bounding boxes and sub-prompts from a long prompt, as shown in Figure \ref{Fig:MLLM}. It is worth noting that our method does not inherently rely on MLLM, as this process can also be performed manually. We provide more experimental analysis in Section \ref{Sec:MLLM}.

\noindent

\section{More Analysis}

\subsection{Effectiveness of Different MLLM}
\label{Sec:MLLM}
To facilitate large-scale evaluation and enhance user-friendliness, we employ MLLMs to automate global region and concept division. Table \ref{Tab:mllm} presents the performance of our framework when integrating different MLLMs as the layout planner. Benefiting from the advanced capabilities of modern MLLMs, the automated division achieves a high success rate (over 92\%, with rare failures primarily attributed to output formatting errors).
Notably, even under planner failure rates of 8\% (Qwen3-VL-7B) and 2\% (GPT-4o), our approach consistently yields substantial overall improvements (↑24\% and ↑25\%, respectively) over the baseline. This demonstrates that our method is highly robust and fundamentally insensitive to occasional planning failures. Furthermore, the comparable gains achieved across various MLLMs—closely approaching the upper bound of GPT-4o with manual improvements (↑28\%)—confirm that the significant performance boost intrinsically stems from our proposed architecture, rather than relying on a specific MLLM.

\begin{table}[h]
\vspace{-4mm}
\resizebox{\linewidth}{!}{
\centering
\begin{tabular}{c|ccccc|c}
\hline
Method                     & Two-object ↑ & Counting ↑ & Position ↑ & Attribute-Binding ↑ & Overall ↑     & Failure Rate \\ \hline
Base-infiniy               & 79.80        & 58.13      & 26.00      & 58.00               & 55.48         & -                    \\ \hline
Qwen3-VL-7B                & 88.73        & 57.88      & 60.50      & 68.00               & 68.76(↑24\%)  & 8\%                  \\
GPT-4o                     & 90.14        & 57.96      & 60.25      & 69.00               & 69.34(↑25\%)  & 2\%                  \\
Qwen3-VL-32B               & 89.63        & 58.07      & 62.00      & 69.00               & 69.53 (↑25\%) & 0\%                  \\
GPT-4o+Manual improvements & 91.92        & 58.25      & 63.00      & 70.00               & 70.79 (↑28\%) & 0\%                  \\ \hline
\end{tabular}
}
\vspace{3mm}
\caption{The impact of different MLLMs on performance. It can be observed that the performance improvement mainly depends on our method, not the MLLM itself (even though current MLLMs can already handle planning content well). Manual improvements refer to prompts that have been further optimized manually (e.g., considering the target size to create the mask).The code we provide in the supp can be run locally without loading any MLLM.}
\label{Tab:mllm}
\vspace{-12mm}
% \vspace{-3mm}
\end{table}

\subsection{More Practical Usages.}
\begin{figure}[h]
    \vspace{-3mm}
    \centering
    \includegraphics[width=\linewidth]{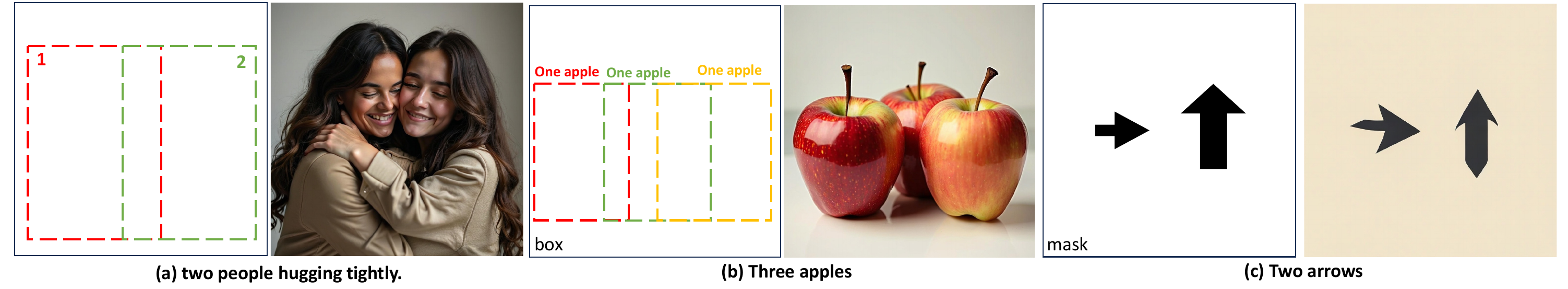}
    \caption{Demonstration of our model's flexibility in practical generation scenarios. (a) Handling overlapping and fine-grained interactive bounding boxes. (b) Seamless generation of multiple instances belonging to the same concept (e.g., three distinct apples) within a single scene. (c) Support for non-rectangular mask guidance to further enhance usability.}
    \label{Fig:practical}
\end{figure}

As shown in Figure~\ref{Fig:practical}(a), our method supports overlapping and fine-grained bounding box specifications. Figure~\ref{Fig:practical}(b) details how we handle prompts with multiple instances of the same concept, treating them as distinct entities in the output (e.g., identifying three separate apples in a scene). Additionally, our model supports non-rectangular masks to improve usability (Figure~\ref{Fig:practical}(c)). Together, these results demonstrate our model's robust flexibility in accommodating multiple instances with diverse spatial layouts and orientations.

\subsection{Different Mask Strategies}
We tried different mask strategies within the receptive field constraint: (a). Const mask: $M=cR$, the cross region is filled with a const value; (b). Guassian mask: defined in Section 3.4 in the paper, use a distance decay weighted mask instead of a const value; (c). TopK-pruning mask: for each query, fill those with top-K highest attention score in the cross region with a const value.
%(topK=2, const value = -100)

\begin{figure}[h]
    \centering
    \includegraphics[width=\linewidth]{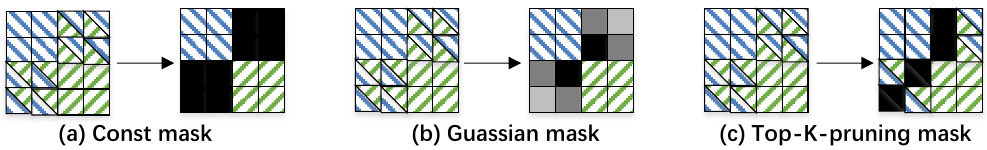}
    \caption{Different mask strategies in receptive field constraints.}
    \label{fig:mask}
    \vspace{-3mm}
\end{figure}

% \begin{table*}[t]
%     \centering
%         \begin{tabular}{c|ccccc}
%         \toprule
%         Methods & Two\_object & Counting & Position & 
%         Attribute\_Binding & Overall \\
%         \midrule
%         const& 87.37&\textbf{59.06}&57.75&69.75&68.48\\
%         topK-pruning&87.37&58.75&59.00&69.25&68.59\\
%        gaussian& \textbf{91.92} & 58.25 & \textbf{63.00} & \textbf{70.00} & \textbf{70.79} \\
%         \bottomrule
%     \end{tabular}
%     \caption{Quantitative experimental results of different mask strategies.}
%     \label{Tab:mask strategy}
% \end{table*}

As shown in Table \ref{Tab:mask strategy}, the Const mask and TopK-pruning mask strategies may ignore the connection in spatial distance, resulting in a decline in performance, while the Gaussian mask calculates the relative spatial proximity of the markers in the early stage of generation and uses moderate interaction to achieve a balance between maintaining regional independence and promoting necessary information transfer, ultimately improving the quality of the generated results.

\begin{table}[h]
    \centering
    \resizebox{\linewidth}{!}{
        \begin{tabular}{c|ccccc}
        \toprule
        Methods & Two\_object & Counting & Position & 
        Attribute\_Binding & Overall \\
        \midrule
        const& 87.37&\textbf{59.06}&57.75&69.75&68.48\\
        topK-pruning&87.37&58.75&59.00&69.25&68.59\\
       gaussian& \textbf{91.92} & 58.25 & \textbf{63.00} & \textbf{70.00} & \textbf{70.79} \\
        \bottomrule
    \end{tabular}
    }
    \vspace{3mm}
    \caption{Quantitative experimental results of different mask strategies.}
    \label{Tab:mask strategy}
    \vspace{-12mm}
\end{table}

% \subsection{Inference Time Cost}
% As shown in Table \ref{tab:time_cost}, we tested the single-image inference time consumption of different components. It can be seen that the inference overhead of Global and Receptive Field Constraints is almost negligible compared to Vanilla VAR. High-Frequency Compensation, due to the frequency domain conversion involved, incurs additional time overhead, but this is traded off for improved aesthetic quality. Since our method does not involve any structural changes, we plan to integrate it with VAR acceleration solutions such as FastVAR \cite{guo2025fastvar} in the future, thereby ensuring further improvements in image quality while avoiding increased inference overhead.

% \begin{table}[h]
%     \centering
%     \resizebox{\linewidth}{!}{
%         \begin{tabular}{c|ccccc}
%         \toprule
%         Methods & Two\_object & Counting & Position & Attribute\_Binding & Overall \\
%         \midrule
%         SynVAR & \textbf{91.92} & \textbf{58.25} & \textbf{63.00} & \textbf{70.00} & \textbf{70.79} 
%         SynVAR & \textbf{91.92} & \textbf{58.25} & \textbf{63.00} & \textbf{70.00} & \textbf{70.79}
%         \bottomrule
%         \end{tabular}
%     }
%     \caption{Quantitative performance of SynVAR with FastVAR on the Geneval.}
%     \label{Tab: Ablation Study}
% \end{table}

\subsection{The compatibility of SynVAR}
We provide the inference overhead of SynVAR compatibility with FastVAR. As shown in Table \ref{Tab: FastVAR Performance}, after being compatible with FastVAR, SynVAR still maintains its superior performance, further demonstrating the strong compatibility of our method.

\begin{table}[h]
    \vspace{-6mm}
    \centering
    \resizebox{\linewidth}{!}{
        \begin{tabular}{c|cccc}
        \toprule
        Methods & Two\_object & Position & Attribute\_Binding & Overall \\
        \midrule
        Fastvar & 79.29  & 27.50 & 60.25 & 56.99 \\
        % \midrule
        %\midrule
        Fastvar + SynVAR & \textbf{88.38}  & \textbf{66.00} & \textbf{65.75} & \textbf{69.16} \\ 
        \bottomrule
        \end{tabular}
    }
    \vspace{3mm}
    \caption{Performance of SynVAR with Fastvar on the Geneval.}
    \label{Tab: FastVAR Performance}
\end{table}

\subsection{More Qualitative Results}
\begin{figure*}[ht]
    \centering
    \includegraphics[width=0.95\linewidth]{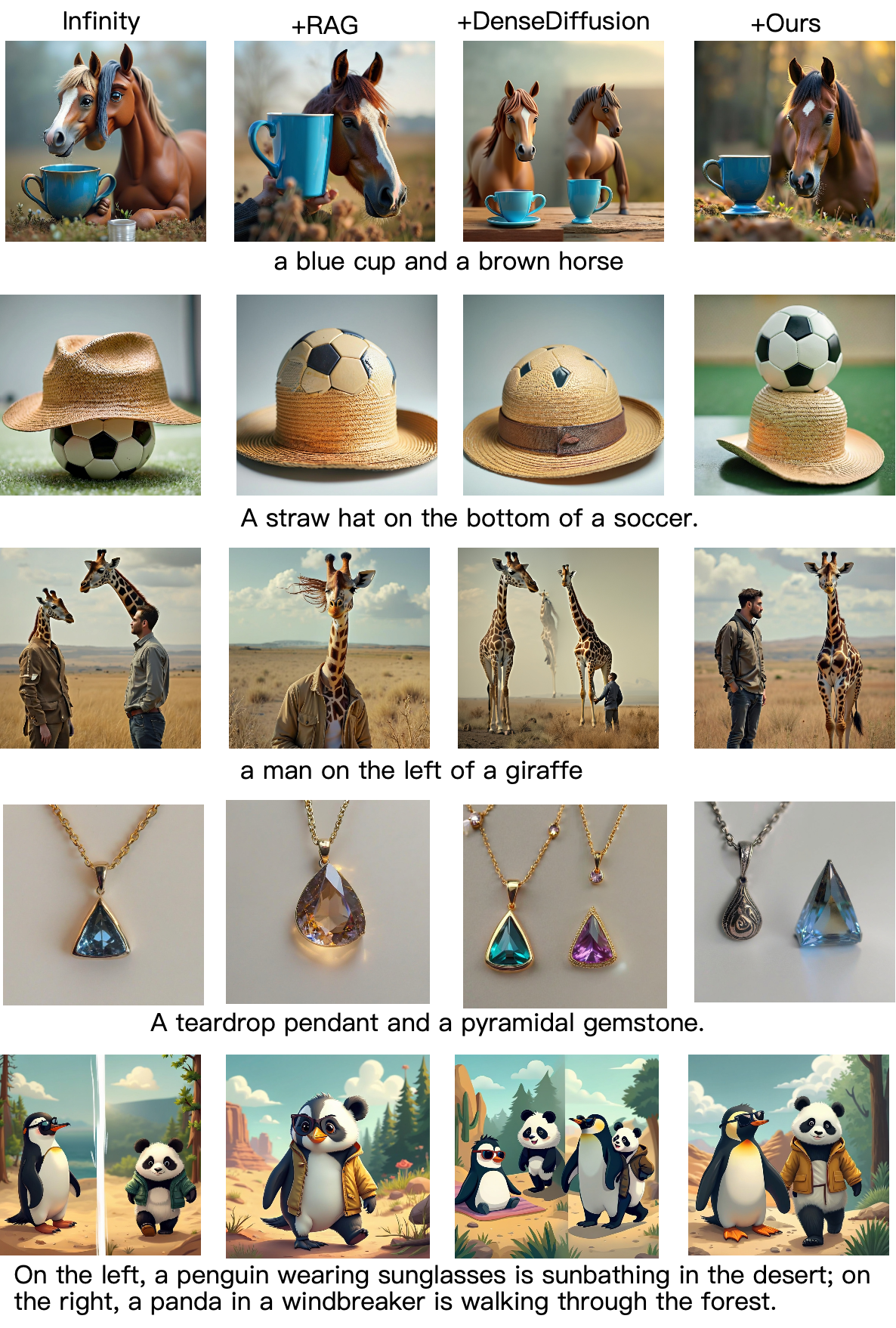}
    \caption{More qualitative results of SynVAR.}
    \label{fig:infinity}
\end{figure*}

\begin{figure*}[ht]
    \centering
    \includegraphics[width=0.98\linewidth]{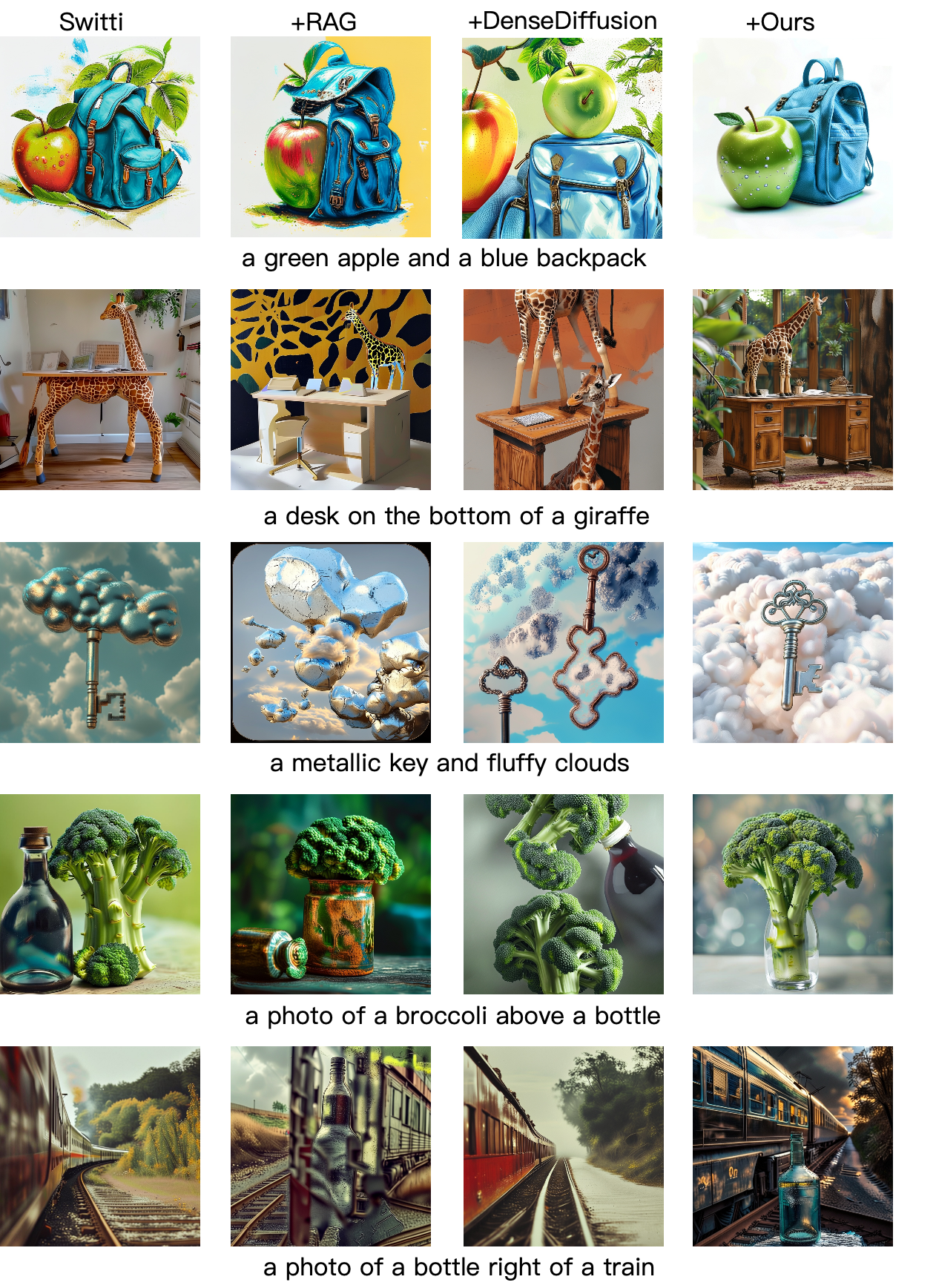}
    \caption{More qualitative results of SynVAR}
    \label{fig:switti}
\end{figure*}

As shown in Figure \ref{fig:infinity} and Figure \ref{fig:switti}, we present additional qualitative results comparing SynVAR, RAG-Diffusion and Densediffusion that further highlighting SynVAR's superior performance in semantic depiction and spatial control. Through the generation of  complex scenes, SynVAR demonstrates its ability to accurately capture the details described in the input text while maintaining overall coherence and naturalness in the generated content.
\subsection{Stability of SynVAR}
\begin{figure*}[ht]
    \centering
    \includegraphics[width=\linewidth]{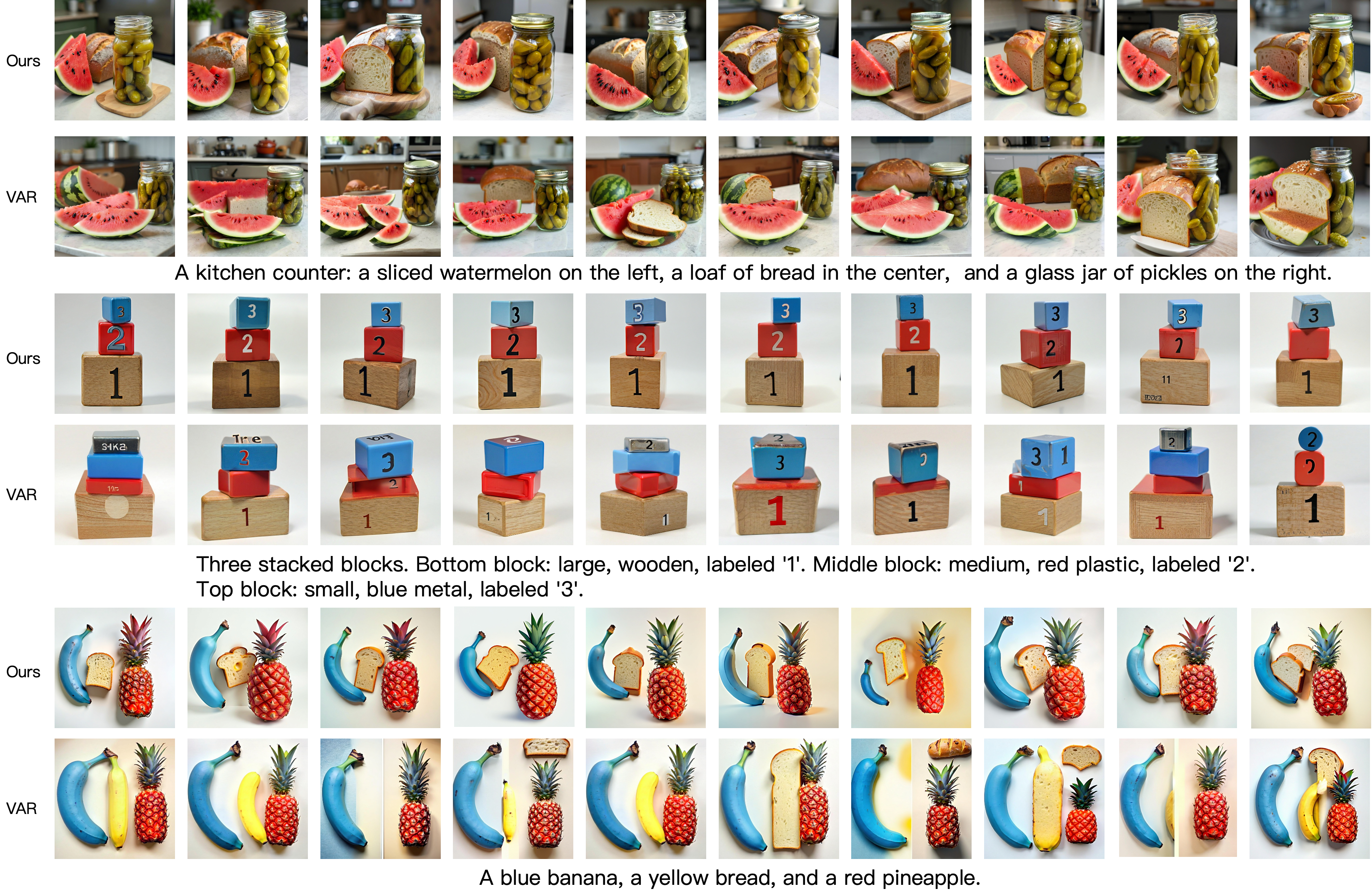}
    \caption{SynVAR generation results using different seeds show more stable generation results.}
    \label{fig:stable}
\end{figure*}

We validated stability by applying different random
seeds under identical parameter settings. As shown in Figure \ref{fig:stable}, occasional suboptimal outcomes were observed, such as incorrect object counts. These cases can be regarded as acceptable variations, given that similar issues are inherent in most existing T2I models during image generation.

% \clearpage
% \bibliography{aaai2026}

% \begin{table}[h]
%     \centering
%     \resizebox{\linewidth}{!}{
%         \begin{tabular}{c|c}
%         \toprule
%         Methods & Two\_object \\
%         \midrule
%         w/o Global Guidance & 74.49 \\
%         % \midrule
%         w/o Receptive Field Constraint & 80.30 \\
%         % \midrule
%         w/o High-Frequency Compensation &88.38 \\ \midrule
%         %\midrule
%         SynVAR & \textbf{91.92} & \textbf{58.25} \\
%         \bottomrule
%         \end{tabular}
%     }
%     \caption{Ablation performance of SynVAR on the Geneval.}
%     \label{Tab: Ablation Study}
% \end{table}

\clearpage

\end{document}